\documentclass{article}

\usepackage{microtype}
\usepackage{graphicx}
\usepackage{subfigure}
\usepackage{booktabs} %

\usepackage[bookmarks=false,linkcolor=blue,urlcolor=blue,colorlinks,citecolor=blue]{hyperref}

\usepackage[preprint]{preprint}

\usepackage{amsmath}
\usepackage{amssymb}
\usepackage{mathtools}
\usepackage{amsthm}
\usepackage{nicefrac}

\usepackage[capitalize,noabbrev]{cleveref}

\theoremstyle{plain}

\theoremstyle{definition}

\theoremstyle{remark}

\usepackage[textsize=tiny]{todonotes}

\newcommand{\x}{{\mathbf x}}

\newcommand{\q}{{\mathbf q}}

\newcommand{\R}{{\mathbb R}}

\newcommand{\norm}[1]{\left\|#1 \right\|}

\DeclareMathOperator*{\argmin}{argmin}
\DeclareMathOperator*{\cost}{cost}

\title{Incrementally-Computable Neural Networks:\\Efficient Inference for Dynamic Inputs}

\author{%
  Or Sharir \\
  Caltech \\
  \texttt{ors@caltech.edu} \\
  \And
  Anima Anandkumar \\
  Caltech \\
  \texttt{anima@caltech.edu} \\
}

\begin{document}

\maketitle

\begin{abstract}
Deep learning often faces the challenge of efficiently processing dynamic inputs, such as sensor data or user inputs.
For example, an AI writing assistant is required to update its suggestions in real time as a document is edited.
Re-running the model each time is expensive, even with compression techniques like knowledge distillation, pruning, or quantization.
Instead, we take an \hbox{\emph{incremental computing}} approach, looking to reuse calculations as the inputs change.
However, the dense connectivity of conventional architectures poses a major obstacle to incremental computation, as even minor input changes cascade through the network and restrict information reuse.
To address this, we use \emph{vector quantization} to discretize intermediate values in the network, which filters out noisy and unnecessary modifications to hidden neurons, facilitating the reuse of their values.
We apply this approach to the transformers architecture, creating an efficient incremental inference algorithm with complexity proportional to the fraction of the modified inputs.
Our experiments with adapting the OPT-125M pre-trained language model demonstrate comparable accuracy on document classification while requiring 12.1X (median) fewer operations for processing sequences of atomic edits.
\end{abstract}

\section{Introduction}
Large language models~(LLMs) based on the transformers architecture~\citep{transformers,BERT,GPT2,GPT3} have revolutionized the entire natural language processing field, achieving remarkable results on a wide range of tasks, including text classification, question answering, document retrieval and many others.
Increasingly, LLMs aid in the writing process itself, analyzing and proposing suggestions as documents are edited~\citep{shi2022effidit,ippolito2022creative}, both in online and offline settings.
In the online case, LLMs must react in real time as a document is edited, word-by-word.
In the offline case, there is a preexisting history of revisions waiting in the queue for processing. 
Even though the difference between each revision is often small~--~just a single word in the online setting~--~today’s LLMs process each revision from scratch, wasting a lot of computational resources.

\begin{figure}[t]
\begin{center}
\centerline{\includegraphics[width=0.85\linewidth]{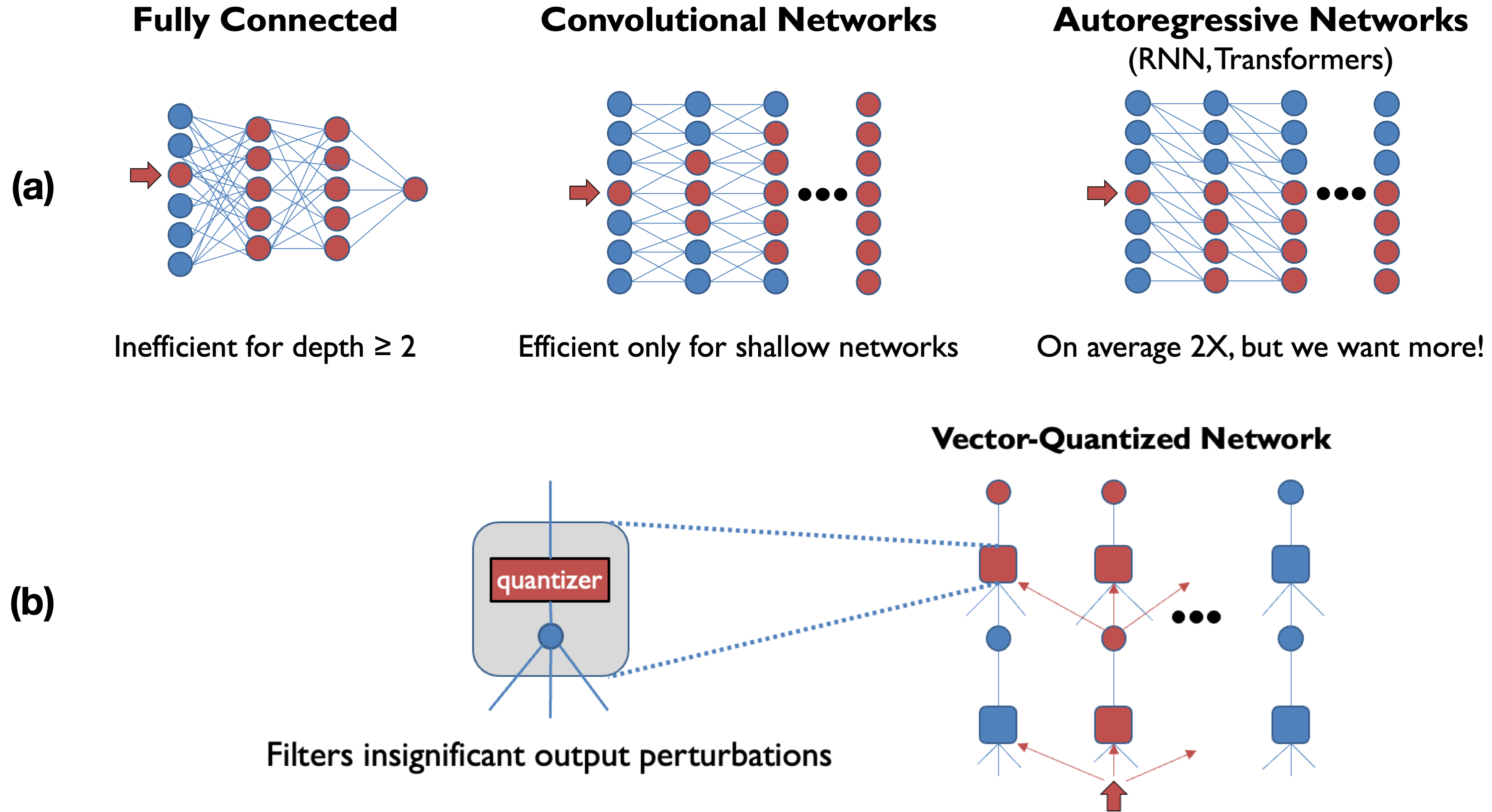}}
\caption{\label{fig:incremental}In conventional architectures \textbf{(a)}, reusing calculations is challenging as modifying even a single input coordinates affects most of the network. We propose vector-quantized networks \textbf{(b)} to filter insignificant output perturbations and enable activation reusability.}
\end{center}
\vskip -0.2in
\end{figure}

An ideal deep-learning model should be able to leverage the similarity between two revisions and reuse information as much as possible.
Such an approach is known as \emph{incremental computing}.
Usually, incremental computing is solved by examining the dependency graph of the computational operations and tracking which nodes are affected by modified inputs.
The prototypical example is a software build system, tracking which source files have changed and recompiling just the files that depend on them.
However, most deep learning architectures, and specifically transformers, are structured such that nearly all neurons depend on nearly all inputs, as illustrated in fig.~\ref{fig:incremental}a.
Having a \emph{densely connected} architecture was proven to be a crucial aspect for producing highly expressive models~\citep{sharir2018expressive,2017benefits,cohen2018analysis}.
But this same property also poses a challenge for reusing calculations. 

We propose to augment transformers with discrete operations, namely, vector quantization~(VQ)~\citep{VQ2017_7a98af17}, to inhibit superfluous effects on the intermediate values in the network.
By only allowing for meaningful effects to propagate, VQ paves the path to incrementally computable neural networks, as illustrated in fig.~\ref{fig:incremental}b.
In sec.~\ref{sec:method}, we describe our complete algorithm for incremental computation of Vector-Quantized Transformers, or VQT for short, with a runtime complexity proportional to the edit distance of the documents.

We demonstrate our approach in sec.~\ref{sec:exp} by adapting the OPT-125M pre-trained language model~\citep{zhang2022opt} and distilling it to its analog VQ-OPT model, followed by fine-tuning on document classification task (IMDB~\citep{maas2011IMDB}), on which it obtains comparable accuracy to the original OPT-125M model.
To measure the runtime performance on processing real-world edits, we scraped Wikipedia edit histories.
Unlike the plain OPT, VQ-OPT requires 12.1X (median) fewer arithmetic operations for online processing of a sequence of atomic edits (replace / insert / delete a single token).
For offline processing of two complete revisions, we observe a more modest 4.7X reduction on average, as expected when modifying larger chunks of the document as opposed to atomic edits.

\section{Related Works}

\textbf{Model compression} is the primary approach for efficient inference.
Knowledge distillation~\citep{10.1145/1150402.1150464,hinton2015distilling,sanh2020distilbert} trains a smaller ``student'' network to imitate the outputs of a larger ``teacher'' network.
Pruning~\citep{han2015learning,han2015deep_compression} and tensor factorizations~\citep{Yu_2017_CVPR,hsu2022language} can be used to find sparse and low-rank representation, respectively, of the network's weights.
Weight and activation quantization~\citep{gholami2021survey} can be used to reduce the memory footprint and communication costs, which can be crucial for running extreme LLMs ($>$1B parameters) on a single GPU or even edge devices.
Combined, they can be very effective, but they do not leverage the vast redundancy found in the editing process.

\textbf{Sparse computation} is a form of selective computation where computation is applied according to some sparse activation mask~\citep{ren2018sbnet,Pan_2018_CVPR,10.1145/3131885.3131906,parger2022deltacnn,chen2023sparsevit}.
Most related to this paper are works that utilize temporal redundancy in video frames.
The general approach is to compute the delta between consecutive activation maps, truncate small values, and then only operate on the remaining sparse values while also trying to account for the loss in numerical precision.
A similar approach was also used for accelerating generative image editing~\citep{li2022efficient}.
While sharing some similarities to our approach, the cited methods focus on a different domain (vision), with different assumptions (gradual pixel change, locality), using a different underlying architecture (convolutional), and resulting in approximate (as opposed to exact) inference, which makes direct comparison nontrivial.
Swapping words can cause abrupt and non-local (though often sparse) effects on the document, very different from the gradual and local nature of video consecutive frames.
Furthermore, our work handles the insertion and deletion of words, akin to shifting pixels around in a non-uniform pattern, while the cited works mostly assume the camera is static.
Nevertheless, future work could investigate the transfer of VQT to the vision domain and allow for a more direct comparison.

\textbf{Vector Quantization}~\citep{ClasssicVQ} has been widely used in machine learning for various means, but never for incremental computing.
Denote with $C\in\R^{q \times d}$ a codebook of $q$ $d$-dimensional vectors, then vector quantization is typically defined as $VQ(\x) = C_{\argmin_i d(\x, C_{i,:}), :}$ for some distance metric $d(\cdot,\cdot)$.
VQ is a discrete operation, both with respect to its parameters, i.e., the codebook, and with respect to its input, meaning its derivatives are zero almost everywhere.
However, several methods have been proposed for defining a ``pseudo-derivative'', e.g., via the straight-through estimator~\citep{VQ2017_7a98af17}, which enables embedding vector quantization into deep learning models and training them with any gradient descent optimizer.
Such methods were employed to define variational auto-encoders with discrete hidden variables~\citep{VQ2017_7a98af17}, learn a  shared latent space for images and text~\citep{Gu_2022_CVPR,yu2022vectorquantized}, compress audio~\citep{zeghidour2021soundstream}, as well as to improve generalization in transformers~\citep{liu2022adaptive,NEURIPS2021_10907813}.
To the best of our knowledge, our work is the first usage of VQ for improving the efficiency at inference time.

\section{Method}\label{sec:method}

To support the incremental computation of transformers, we propose two simple modifications to the self-attention layer in the transformers architecture: (i) Append a VQ layer the self-attention outputs the outputs, and (ii) replace the softmax normalization of the attention matrix with any element-wise non-linearity.
This amounts to the following vector-quantized self-attention operation:
\begin{align}\label{eq:vq-attn}
	O &= \mathrm{VQ}(\sigma(QK^T)V),
\end{align}
where $Q, K,$ and $V$ are the query, key, and value matrices, respectively, $\sigma$ is the non-linear element-wise activation (e.g., a GELU), and $\mathrm{VQ}$ the VQ module applied on every row separately.
For multi-head self-attention, the VQ layer takes as input the concatenation of all attention heads, followed by an additional linear layer to mix the heads.
It was recently proposed~\cite{pmlr-v162-hua22a,ma2022mega} that using such an element-wise non-linearity has comparable performance to softmax while being easier to optimize on the hardware level.
Here, we employ it for numerical stability, because though reusing calculations with softmax is mathematically possible, numerically it can lead to a significant loss of precision.

Using the above module, we can now present how we can reuse calculations.
As discussed in the introductions, there are two settings of interest, online editing, operating on one small change at a time, and offline batch processing, operating on a batch of revisions.
To simplify the presentation, we will present our algorithm for the offline setting, where the online setting is equivalent to a batch of size 2 and keeping a cache for the first input.
Furthermore, we first assume the only valid edit operations are token replacements, and will handle the token insertion and deletion in subsection~\ref{sec:insert}.

We reuse calculations in two steps.
First, we describe a compressed format for storing the quantized activations, leveraging the redundancy in the quantized intermediate values of similar revisions.
Second, we show how this compressed format can be directly manipulated to efficiently implement any operation in the VQ Transformer.

\subsection{Compressing Vector-Quantized Activations}

\begin{figure*}[t]
\begin{center}
\centerline{\includegraphics[width=0.9\linewidth]{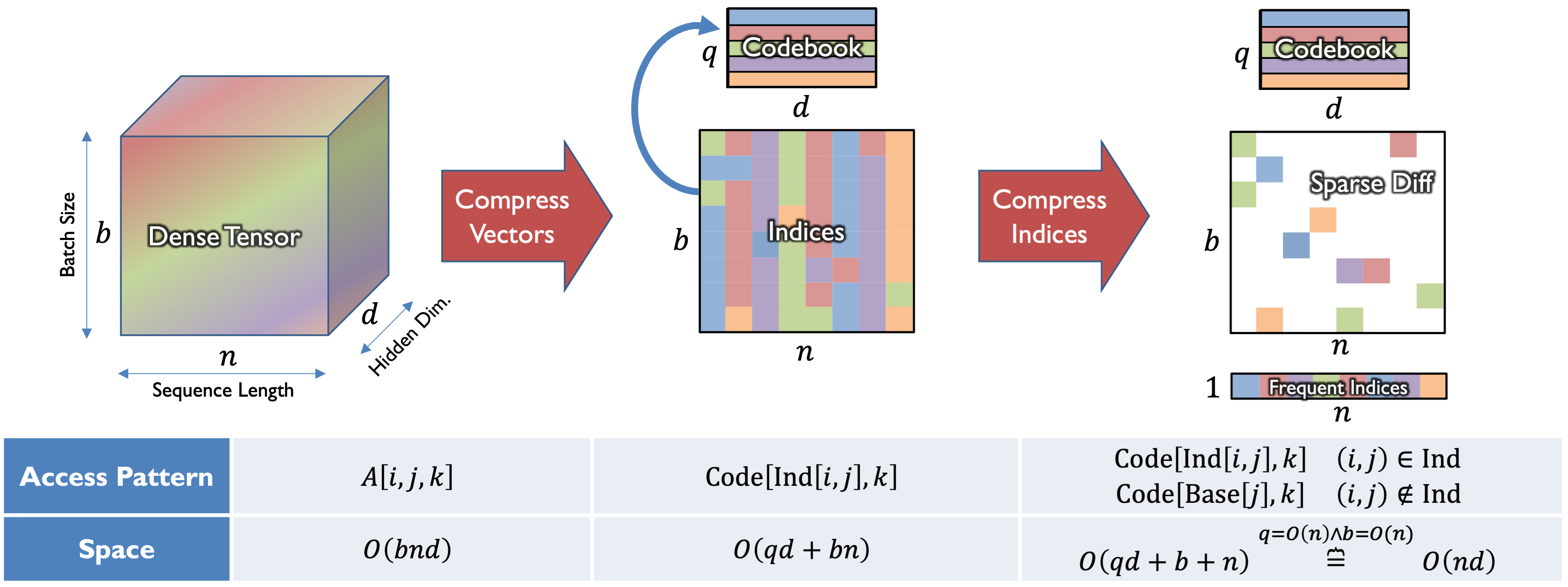}}
\caption{\label{fig:compression}Illustration of the compression scheme of the vector-quantized activations through our VQ-Transformer network.}
\end{center}
\vskip -0.2in
\end{figure*}

The intermediate values in the transformer can typically be described as a 3D tensor $X \in \R^{b \times n \times d}$, where $b, n, $ and $d$ denoted the batch size, sequence length, and hidden dimension, respectively.
Since we assume $X$ is vector-quantized\footnote{For the initial token embeddings, since they are the result of fetching vectors from an embedding matrix, they too can be considered vector-quantized.}, each hidden vector in any batch and sequence location is matched against some codebook matrix $C \in \R^{q \times d}$, represent the unique vectors found in $X$.
This means we can represent it more compactly with an indices matrix $P \in \{1, \ldots, q\}^{b \times n}$ pointing to the codebook vectors, i.e., $X_{ijk} = C_{P_{ij}k}$.
We emphasize that $C$ is not the same as the codebook in the VQ layer itself, but rather the representation of the unique sets of vectors in $X$.
It could have both fewer vectors than in VQ as well as more, depending on the sequence of operations up to this point.

If we also assume the edit distance between revisions is small, i.e., few tokens were replaced per revision, then the indices in $P$ should agree on most sequence locations across the batch dimension.
This has two implications. 
First, $q = O(n + b)$ since the number of unique indices cannot be more than the length of any row plus the few indices that are different per row.
Second, at any sequence location across the batch dimension, most indices should be the same, meaning we can further compress $P$ through a sparse representation.
Namely, at any sequence location pick the most frequent index, resulting in a base set of indices per location, and then store only the coordinates in $P$ which differ from the base.
This sparse representation has an $O(n + b)$ storage complexity.
In total, we get an $O((n + b)d)$ storage complexity, and since typically $b = O(n)$ as well, it reduces to just $O(nd)$ down from $O(bnd)$ for the dense representation of $X$.
The entire compression scheme is illustrated in fig.~\ref{fig:compression}.

\subsection{Efficient Operations Over the Compressed Format}

Having a compressed representation only helps if we can efficiently operate over it.
The most common type of operation in transformers can be broadly described as a \emph{identical per-location vector operation}, i.e., any vector-valued operation that is applied on each location separately.
Formally, they can be written as $Y = F(X)$ such that $Y_{ij:} = f(X_{ij:})$ for some $f:\R^d \to \R^d$, and it has an $O(bn \cost(f))$ complexity.
This includes normalization layers, linear layers, scaling by a constant, activation functions, and others.
In most common configurations, they together constitute more than 70\% of the total floating-point operations in the forward pass, and for extreme models like GPT3~\citep{GPT3} it is over 97\%.
Fortunately, we map this operation very efficiently to our compressed format.
If $X$ is represented by $(P, C)$, following the same notations as before, then $Y$ is equivalent to $(P, F(C))$, i.e., operating only on the codebook matrix while keeping the indices unchanged.
This relation holds because
\begin{align}
	Y_{ij:} = f(X_{ij:}) = f(C_{P_{ij}:} = f(C)_{P_{ij}:} .
\end{align}
Thus, the complexity of applying per-location operations on our compressed format is mere $O(q \cost(f)) = O(n \cost(f))$, i.e., reducing the dependency on the batch size.

For other operations not covered by the above, e.g., self-attention and residual connections, we can show similar improved efficiency, but we defer their description to app.~\ref{app:other-ops}.
In the same appendix, we also explain how the cost of the VQ layer itself could be reduced during inference.
In total, we can show that any batch operation in the network can be computed with an equivalent complexity to processing a single document (up to constants).
Of course, this analysis depends on the assumption that the VQ layer can indeed converge to a solution during training that maximizes the redundancy found in the input.
We demonstrate in our experiments in sec.~\ref{sec:exp} that indeed this method can attain significant computational savings in practice.

\subsection{Supporting Token Insertion and Deletion}\label{sec:insert}

Previously, we assumed for simplicity that the only valid edit operations are replacing tokens, keeping the total number of tokens in a document fixed.
The main issue with inserting or deleting tokens is the effect it has on the positional embeddings, effectively shifting them around the edit locations.
If we were to use conventional positional embeddings, then inserting a token would lead to nearly all token representations to be different between two revisions, and so little opportunity for reusability.

Our solution is training with a \emph{sampled} absolute positional embedding.
During training, instead of using a contiguous range of positional indices~\citep{transformers}, we sample per document a random ordered subset of the maximum number of positional embeddings.
So instead of always assigning positions $1,2,3,4$ to the first 4 tokens, we sample $p_1 < p_2 < p_3 < p_4$, and use those values to fetch their respective vectors from the positional embedding matrix.

In essence, we are training the absolute positional embedding to be relational, forcing the network to only leverage the order between positions and not their absolute value.
The benefit of such a scheme is that we can now spread the initial positions, allowing for gaps in between tokens, so new tokens could be inserted with suitable positions.
We can further use the attention mask matrix to ensure other tokens do not attend to the pad locations, thus ensuring that each prediction is self-consistent and independent of the other revision in the batch.

For the online setting, every so often might require reindexing the positional values after the initial gaps are filled, akin to defragmentation.
As long as it does not happen too often, the expected speedup will remain effective.
To ensure this is the case, we can use a very large pool of positional embedding vectors, e.g., on the order of 100X more than the maximal sequence length.
It is reasonable to train with such a large number of positional vectors due to the well-known coupon collector problem.
See app.~\ref{app:coupon} for further discussion.

For the offline batch setting, we do not need as much overhead.
We can simply align all revisions by introducing pad tokens for missing or misaligned tokens, and then assign a contiguous range of positions according to the longest document including padding.

\section{Experiments}\label{sec:exp}

We demonstrate our method by training a decoder transformer model following the OPT~\citep{zhang2022opt} family of models.
Since training large language models from scratch is expensive, we opt instead to use knowledge distillation to adapt the base OPT 125M parameters model to fit our architecture.
We follow the procedure of \citet{sanh2020distilbert}. We keep all the parameters of the original model while adding the VQ layers in each self-attention module and changing softmax to GELU~\citep{GELU}.

We specifically use multi-head VQ, i.e., each vector is split into smaller chunks and each is matched against a codebook of 64 vectors, thus the effective codebook is of size $64^{\textrm{heads}}$.
For the pseudo-gradient, we use a variant of the Gumbel-Softmax estimator~\citep{jang2017categorical}.

During training, we use the sampled positional embedding scheme of sec.~\ref{sec:insert}, where we initialize the larger positional embedding matrix from the original by simply repeating each positional embedding to match the new size.

For distillation, we use the Pile~\citep{pile}, which constitutes a large subset of the corpus used to train OPT.
Unlike \citet{zhang2022opt}, we used the raw corpus without extra filtering and deduplication steps.
We trained for 100K iterations, using 16384 tokens per iteration with a max sequence length of 2048, 5K iterations of linear warm-up to a learning rate of 5e-4, followed by a cosine decay to 5e-5.

As an alternative to our incremental computing approach, we also distilled OPT to a smaller OPT model with 6 instead of 12 layers, following the same initialization as \citet{sanh2020distilbert} and training using the same method as above (though \citet{sanh2020distilbert} trained their distilled models for longer). 

\begin{table}[t]
\caption{Accuracy and F1 on the IMDB document classification dataset. $h$ denotes the number of heads in each VQ layer.}
\label{table:accuracy}
\begin{center}
\begin{tabular}{lccc}
\toprule
\textbf{Model}&\multicolumn{2}{c}{\textbf{IMDB}} \\
\midrule
              & Accuracy & F1   \\
\midrule
RoBERTa       & 95.3     & 95.0  \\
OPT-125M      & 94.4     & 94.5  \\
DistilOPT     & 92.4     & 92.3  \\
VQ-OPT (h=2)  & 90.3     & 90.4  \\
VQ-OPT (h=4)  & 91.6     & 91.6  \\
\bottomrule
\end{tabular}
\end{center}
\end{table}

\begin{table}[t]
\caption{Theoretical speedups for processing sequence of edits to documents, based on measuring relative reduction in arithmetic operations on subset of 500 random edits (either atomic or whole revisions) scraped from English Wikipedia.}
\label{table:speedups}
\begin{center}
\begin{tabular}{lccc}
\toprule
\textbf{Model} & Atomic & Entire Revision & First 5\%    \\
\midrule
OPT-125M      & 1X      & 1X    &  1X \\
DistilOPT     & 2X      & 2X    &  2X \\
VQ-OPT (h=2)  & 12.1X   & 4.7X  &  4.8X \\
VQ-OPT (h=4)  & 5.2X    & 2.5X  &  2.2X \\
\bottomrule
\end{tabular}
\end{center}
\end{table}

After the initial distillation phase, we fine-tuned our VQ-OPT model, the original OPT, and our DistilOPT model on sentiment analysis using the IMDB reviews dataset.
It was chosen as it operates over relatively long documents, and as it could represent tasks relevant to a hypothetical writing assistant.
We present the results in table~\ref{table:accuracy}, and we included the reported results of RoBERTa~\citep{liu2019roberta,ding-etal-2021-ernie} as a reference.
As can be seen, both VQ-OPT and DistilOPT attain comparable results to the original OPT, demonstrating that our proposed approach can retain most of the capacity of the original OPT, with VQ-OPT retaining 95-97\% of the accuracy of OPT-125M, depending on the number of VQ heads (i.e., codebook size).

\begin{figure}[t]
\begin{center}
\centerline{\includegraphics[width=0.6\linewidth]{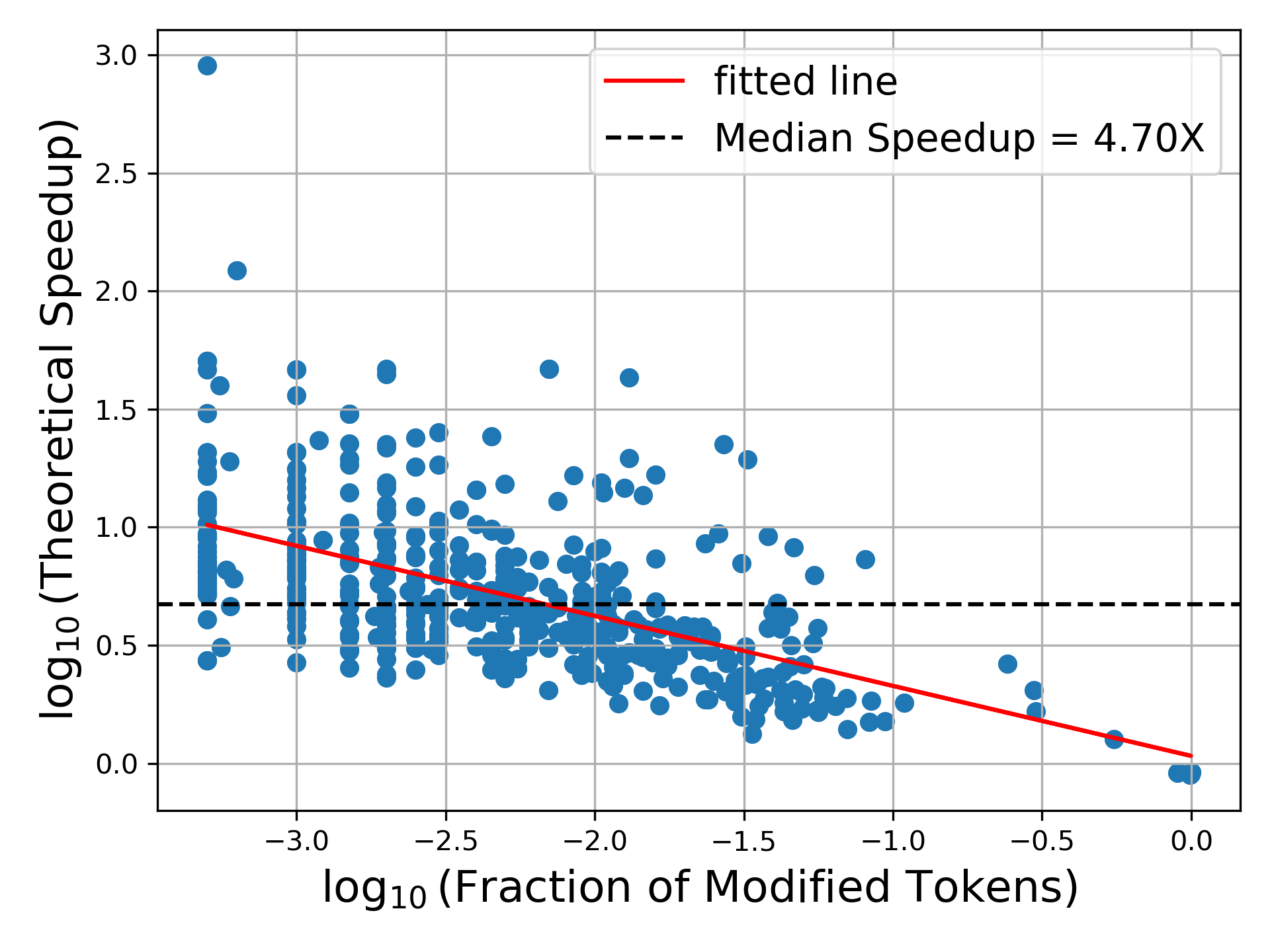}}
\caption{\label{fig:offline}Relative reduction in arithmetic operations for offline processing of two complete revisions. Each point corresponds to 2 consecutive revisions in a Wikipedia edit history with respect to the fraction of modified tokens, i.e., edit distance, demonstrating complexity is proportional to edit distance. 4.7X is the median reduction. }
\end{center}
\vskip -0.2in
\end{figure}

\begin{figure}[t]
\begin{center}
\centerline{\includegraphics[width=0.6\linewidth]{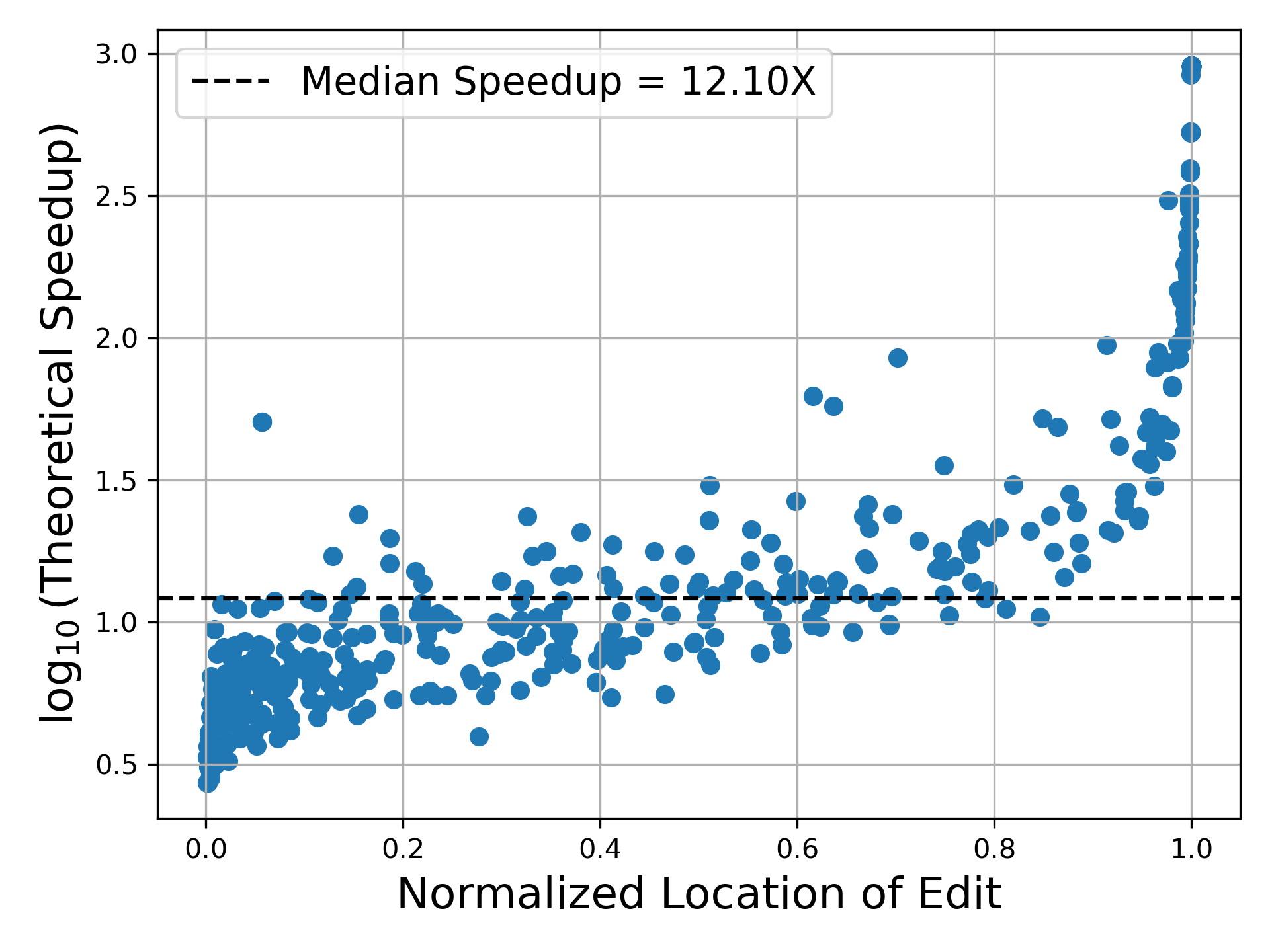}}
\caption{\label{fig:online}Relative reduction in arithmetic operations for online processing of atomic edits (logarithmic scale). Each point corresponds to the processing of a single atomic edit at a given normalized location. 12.1X is the median reduction in cost.}
\end{center}
\vskip -0.2in
\end{figure}

Finally, we measure the runtime complexity for processing edits in a document.
We examine the offline case of processing an entirely new revision of a document and the online case of atomic edits.
We also examine the case of atomic edits happening in the first 5\% of tokens for a more fair comparison to OPT-125M and DistilOPT which do not explicitly leverage the redundancy found in edited documents. 
For every case, we have collected 500 pairs of consecutive revisions from the edit history of Wikipedia articles.
The articles were selected from the list of featured articles that also had a long history of revisions, and the specific revisions were randomly sampled from the set of revisions having at least 1536 tokens and no more than 2048.
We chose revisions of specific lengths to ensure they will fit within all architectures as well as reduce the effect of variable sequence length on the complexity estimation.
Revisions with purely metadata changes, as well as revisions that were later reverted (often due to malicious editing), were pruned from the histories.

We focus on the VQ-OPT with 2 heads, as it retains as much as 95\% of the accuracy of OPT-125M while providing the most computational benefits.
We measured the theoretical arithmetic operations required for the forward pass assuming the previous revision was already processed.
We present the ratio of  arithmetic operations for the original OPT to VQ-OPT as the theoretical speedup under ideal implementation.
For the offline case where we process two complete revisions, we observe that on average the theoretical speedup is 4X. However, a closer examination reveals that the theoretical speedup is simply proportional to the fraction of the modified tokens, i.e., changing fewer words results in more opportunity to reuse calculations, as presented in figure~\ref{fig:offline}.

To simulate the online case where a user edits a document word-by-word, we pick a random modified location in a given pair of revisions and keep all the changes up to that point in the document and ignore all that came after it.
Here we see that the median theoretical speedup is 12.1X, and that there is a correlation between the relative location of the edit and the theoretical speedup.
Our results for the online case are presented in figure~\ref{fig:online}.

In both the online and offline cases it is important to keep in mind that for DistilOPT is limited to just a 2X improvement over OPT-125M. 

\section{Discussion}

We proposed the first incremental computable transformer model, attaining significant computational savings while performing comparably to its target pre-trained model.

Nevertheless, we wish the emphasize some of the limitations of these early findings.
It is well known that as language models grow, their behavior can change in a non-linear way~\citep{GPT3}.
Namely, it is well-documented that post-training quantization of very large language models ($\ge$6B) is more difficult due to the presence of activation outliers~\citep{xiao2022smoothquant}.
Testing our method on such scales could reveal similar challenges, though some of the proposed solutions could be applicable to our case.
Furthermore, while the fundamental assumption of our approach is that minor input modifications should not typically result in massive modifications in the computational graph, it is trivial to imagine pathological edge cases where that assumption would not hold.
It is thus crucial to further experiment on a wide variety of NLP tasks and assess the runtime complexity per task.

Regardless of the above limitations, incremental computation sets a completely new path to efficient inference, which was under-explored in the machine learning community.
While this paper has focused on its application to natural language, the same method is highly relevant to many other domains, including different modalities (video, genes), as well as settings (e.g., agent-environment interaction loop).
We plan to explore these avenues in future works.

\section*{Acknowledgments}
The authors would like to thank Devin Chotzen-Hartzell for his contributions to preliminary experiments and initial code for scraping Wikipedia edit histories.

\bibliography{refs}

\begin{thebibliography}{40}
\providecommand{\natexlab}[1]{#1}
\providecommand{\url}[1]{\texttt{#1}}
\expandafter\ifx\csname urlstyle\endcsname\relax
  \providecommand{\doi}[1]{doi: #1}\else
  \providecommand{\doi}{doi: \begingroup \urlstyle{rm}\Url}\fi

\bibitem[Brown et~al.(2020)Brown, Mann, Ryder, Subbiah, Kaplan, Dhariwal,
  Neelakantan, Shyam, Sastry, Askell, Agarwal, Herbert-Voss, Krueger, Henighan,
  Child, Ramesh, Ziegler, Wu, Winter, Hesse, Chen, Sigler, Litwin, Gray, Chess,
  Clark, Berner, McCandlish, Radford, Sutskever, and Amodei]{GPT3}
Brown, T., Mann, B., Ryder, N., Subbiah, M., Kaplan, J.~D., Dhariwal, P.,
  Neelakantan, A., Shyam, P., Sastry, G., Askell, A., Agarwal, S.,
  Herbert-Voss, A., Krueger, G., Henighan, T., Child, R., Ramesh, A., Ziegler,
  D., Wu, J., Winter, C., Hesse, C., Chen, M., Sigler, E., Litwin, M., Gray,
  S., Chess, B., Clark, J., Berner, C., McCandlish, S., Radford, A., Sutskever,
  I., and Amodei, D.
\newblock Language models are few-shot learners.
\newblock In Larochelle, H., Ranzato, M., Hadsell, R., Balcan, M., and Lin, H.
  (eds.), \emph{Advances in Neural Information Processing Systems}, volume~33,
  pp.\  1877--1901. Curran Associates, Inc., 2020.
\newblock URL
  \url{https://proceedings.neurips.cc/paper_files/paper/2020/file/1457c0d6bfcb4967418bfb8ac142f64a-Paper.pdf}.

\bibitem[Buciluundefined et~al.(2006)Buciluundefined, Caruana, and
  Niculescu-Mizil]{10.1145/1150402.1150464}
Buciluundefined, C., Caruana, R., and Niculescu-Mizil, A.
\newblock Model compression.
\newblock In \emph{Proceedings of the 12th ACM SIGKDD International Conference
  on Knowledge Discovery and Data Mining}, KDD '06, pp.\  535–541, New York,
  NY, USA, 2006. Association for Computing Machinery.
\newblock ISBN 1595933395.
\newblock \doi{10.1145/1150402.1150464}.
\newblock URL \url{https://doi.org/10.1145/1150402.1150464}.

\bibitem[Cavigelli et~al.(2017)Cavigelli, Degen, and
  Benini]{10.1145/3131885.3131906}
Cavigelli, L., Degen, P., and Benini, L.
\newblock Cbinfer: Change-based inference for convolutional neural networks on
  video data.
\newblock In \emph{Proceedings of the 11th International Conference on
  Distributed Smart Cameras}, ICDSC 2017, pp.\  1–8, New York, NY, USA, 2017.
  Association for Computing Machinery.
\newblock ISBN 9781450354875.
\newblock \doi{10.1145/3131885.3131906}.
\newblock URL \url{https://doi.org/10.1145/3131885.3131906}.

\bibitem[Chen et~al.(2023)Chen, Liu, Tang, Yi, Zhao, and
  Han]{chen2023sparsevit}
Chen, X., Liu, Z., Tang, H., Yi, L., Zhao, H., and Han, S.
\newblock Sparsevit: Revisiting activation sparsity for efficient
  high-resolution vision transformer.
\newblock In \emph{IEEE/CVF Conference on Computer Vision and Pattern
  Recognition (CVPR)}, 2023.

\bibitem[Cohen et~al.(2018)Cohen, Sharir, Levine, Tamari, Yakira, and
  Shashua]{cohen2018analysis}
Cohen, N., Sharir, O., Levine, Y., Tamari, R., Yakira, D., and Shashua, A.
\newblock Analysis and design of convolutional networks via hierarchical tensor
  decompositions, 2018.

\bibitem[Devlin et~al.(2019)Devlin, Chang, Lee, and Toutanova]{BERT}
Devlin, J., Chang, M.-W., Lee, K., and Toutanova, K.
\newblock {BERT}: Pre-training of deep bidirectional transformers for language
  understanding.
\newblock In \emph{Proceedings of the 2019 Conference of the North {A}merican
  Chapter of the Association for Computational Linguistics: Human Language
  Technologies, Volume 1 (Long and Short Papers)}, pp.\  4171--4186,
  Minneapolis, Minnesota, June 2019. Association for Computational Linguistics.
\newblock \doi{10.18653/v1/N19-1423}.
\newblock URL \url{https://aclanthology.org/N19-1423}.

\bibitem[Ding et~al.(2021)Ding, Shang, Wang, Sun, Tian, Wu, and
  Wang]{ding-etal-2021-ernie}
Ding, S., Shang, J., Wang, S., Sun, Y., Tian, H., Wu, H., and Wang, H.
\newblock {ERNIE}-{D}oc: A retrospective long-document modeling transformer.
\newblock In \emph{Proceedings of the 59th Annual Meeting of the Association
  for Computational Linguistics and the 11th International Joint Conference on
  Natural Language Processing (Volume 1: Long Papers)}, pp.\  2914--2927,
  Online, August 2021. Association for Computational Linguistics.
\newblock \doi{10.18653/v1/2021.acl-long.227}.
\newblock URL \url{https://aclanthology.org/2021.acl-long.227}.

\bibitem[Gao et~al.(2020)Gao, Biderman, Black, Golding, Hoppe, Foster, Phang,
  He, Thite, Nabeshima, Presser, and Leahy]{pile}
Gao, L., Biderman, S., Black, S., Golding, L., Hoppe, T., Foster, C., Phang,
  J., He, H., Thite, A., Nabeshima, N., Presser, S., and Leahy, C.
\newblock The {P}ile: An 800gb dataset of diverse text for language modeling.
\newblock \emph{arXiv preprint arXiv:2101.00027}, 2020.

\bibitem[Gholami et~al.(2021)Gholami, Kim, Dong, Yao, Mahoney, and
  Keutzer]{gholami2021survey}
Gholami, A., Kim, S., Dong, Z., Yao, Z., Mahoney, M.~W., and Keutzer, K.
\newblock A survey of quantization methods for efficient neural network
  inference, 2021.

\bibitem[Gray(1984)]{ClasssicVQ}
Gray, R.
\newblock Vector quantization.
\newblock \emph{IEEE ASSP Magazine}, 1\penalty0 (2):\penalty0 4--29, 1984.
\newblock \doi{10.1109/MASSP.1984.1162229}.

\bibitem[Gu et~al.(2022)Gu, Chen, Bao, Wen, Zhang, Chen, Yuan, and
  Guo]{Gu_2022_CVPR}
Gu, S., Chen, D., Bao, J., Wen, F., Zhang, B., Chen, D., Yuan, L., and Guo, B.
\newblock Vector quantized diffusion model for text-to-image synthesis.
\newblock In \emph{Proceedings of the IEEE/CVF Conference on Computer Vision
  and Pattern Recognition (CVPR)}, pp.\  10696--10706, June 2022.

\bibitem[Han et~al.(2015)Han, Pool, Tran, and Dally]{han2015learning}
Han, S., Pool, J., Tran, J., and Dally, W.
\newblock Learning both weights and connections for efficient neural network.
\newblock In \emph{Advances in Neural Information Processing Systems (NIPS)},
  pp.\  1135--1143, 2015.

\bibitem[Han et~al.(2016)Han, Mao, and Dally]{han2015deep_compression}
Han, S., Mao, H., and Dally, W.~J.
\newblock Deep compression: Compressing deep neural networks with pruning,
  trained quantization and huffman coding.
\newblock \emph{International Conference on Learning Representations (ICLR)},
  2016.

\bibitem[Hendrycks \& Gimpel(2020)Hendrycks and Gimpel]{GELU}
Hendrycks, D. and Gimpel, K.
\newblock Gaussian error linear units (gelus), 2020.

\bibitem[Hinton et~al.(2015)Hinton, Vinyals, and Dean]{hinton2015distilling}
Hinton, G., Vinyals, O., and Dean, J.
\newblock Distilling the knowledge in a neural network, 2015.

\bibitem[Hsu et~al.(2022)Hsu, Hua, Chang, Lou, Shen, and Jin]{hsu2022language}
Hsu, Y.-C., Hua, T., Chang, S., Lou, Q., Shen, Y., and Jin, H.
\newblock Language model compression with weighted low-rank factorization.
\newblock In \emph{International Conference on Learning Representations}, 2022.
\newblock URL \url{https://openreview.net/forum?id=uPv9Y3gmAI5}.

\bibitem[Hua et~al.(2022)Hua, Dai, Liu, and Le]{pmlr-v162-hua22a}
Hua, W., Dai, Z., Liu, H., and Le, Q.
\newblock Transformer quality in linear time.
\newblock In Chaudhuri, K., Jegelka, S., Song, L., Szepesvari, C., Niu, G., and
  Sabato, S. (eds.), \emph{Proceedings of the 39th International Conference on
  Machine Learning}, volume 162 of \emph{Proceedings of Machine Learning
  Research}, pp.\  9099--9117. PMLR, 17--23 Jul 2022.
\newblock URL \url{https://proceedings.mlr.press/v162/hua22a.html}.

\bibitem[Ippolito et~al.(2022)Ippolito, Yuan, Coenen, and
  Burnam]{ippolito2022creative}
Ippolito, D., Yuan, A., Coenen, A., and Burnam, S.
\newblock Creative writing with an ai-powered writing assistant: Perspectives
  from professional writers, 2022.

\bibitem[Jang et~al.(2017)Jang, Gu, and Poole]{jang2017categorical}
Jang, E., Gu, S., and Poole, B.
\newblock Categorical reparameterization with gumbel-softmax.
\newblock In \emph{International Conference on Learning Representations}, 2017.
\newblock URL \url{https://openreview.net/forum?id=rkE3y85ee}.

\bibitem[Levine et~al.(2017)Levine, Sharir, and Shashua]{2017benefits}
Levine, Y., Sharir, O., and Shashua, A.
\newblock Benefits of depth for long-term memory of recurrent networks.
\newblock \emph{arXiv preprint arXiv:1710.09431}, 2017.

\bibitem[Li et~al.(2022)Li, Lin, Meng, Ermon, song han, and
  Zhu]{li2022efficient}
Li, M., Lin, J., Meng, C., Ermon, S., song han, and Zhu, J.-Y.
\newblock Efficient spatially sparse inference for conditional {GAN}s and
  diffusion models.
\newblock In Oh, A.~H., Agarwal, A., Belgrave, D., and Cho, K. (eds.),
  \emph{Advances in Neural Information Processing Systems}, 2022.
\newblock URL \url{https://openreview.net/forum?id=AUz5Oig77OS}.

\bibitem[Liu et~al.(2021)Liu, Lamb, Kawaguchi, ALIAS PARTH~GOYAL, Sun, Mozer,
  and Bengio]{NEURIPS2021_10907813}
Liu, D., Lamb, A.~M., Kawaguchi, K., ALIAS PARTH~GOYAL, A.~G., Sun, C., Mozer,
  M.~C., and Bengio, Y.
\newblock Discrete-valued neural communication.
\newblock In Ranzato, M., Beygelzimer, A., Dauphin, Y., Liang, P., and Vaughan,
  J.~W. (eds.), \emph{Advances in Neural Information Processing Systems},
  volume~34, pp.\  2109--2121. Curran Associates, Inc., 2021.
\newblock URL
  \url{https://proceedings.neurips.cc/paper_files/paper/2021/file/10907813b97e249163587e6246612e21-Paper.pdf}.

\bibitem[Liu et~al.(2022)Liu, Lamb, Ji, Notsawo, Mozer, Bengio, and
  Kawaguchi]{liu2022adaptive}
Liu, D., Lamb, A., Ji, X., Notsawo, P., Mozer, M., Bengio, Y., and Kawaguchi,
  K.
\newblock Adaptive discrete communication bottlenecks with dynamic vector
  quantization, 2022.

\bibitem[Liu et~al.(2019)Liu, Ott, Goyal, Du, Joshi, Chen, Levy, Lewis,
  Zettlemoyer, and Stoyanov]{liu2019roberta}
Liu, Y., Ott, M., Goyal, N., Du, J., Joshi, M., Chen, D., Levy, O., Lewis, M.,
  Zettlemoyer, L., and Stoyanov, V.
\newblock Roberta: A robustly optimized bert pretraining approach.
\newblock \emph{arXiv preprint arXiv:1907.11692}, 2019.

\bibitem[Ma et~al.(2022)Ma, Zhou, Kong, He, Gui, Neubig, May, and
  Luke]{ma2022mega}
Ma, X., Zhou, C., Kong, X., He, J., Gui, L., Neubig, G., May, J., and Luke, Z.
\newblock Mega: Moving average equipped gated attention.
\newblock \emph{arXiv preprint arXiv:2209.10655}, 2022.

\bibitem[Maas et~al.(2011)Maas, Daly, Pham, Huang, Ng, and Potts]{maas2011IMDB}
Maas, A.~L., Daly, R.~E., Pham, P.~T., Huang, D., Ng, A.~Y., and Potts, C.
\newblock Learning word vectors for sentiment analysis.
\newblock In \emph{Proceedings of the 49th Annual Meeting of the Association
  for Computational Linguistics: Human Language Technologies}, pp.\  142--150,
  Portland, Oregon, USA, June 2011. Association for Computational Linguistics.
\newblock URL \url{https://aclanthology.org/P11-1015}.

\bibitem[Pan et~al.(2018)Pan, Lin, Fang, Huang, Zhou, and Lu]{Pan_2018_CVPR}
Pan, B., Lin, W., Fang, X., Huang, C., Zhou, B., and Lu, C.
\newblock Recurrent residual module for fast inference in videos.
\newblock In \emph{Proceedings of the IEEE Conference on Computer Vision and
  Pattern Recognition (CVPR)}, June 2018.

\bibitem[Parger et~al.(2022)Parger, Tang, Twigg, Keskin, Wang, and
  Steinberger]{parger2022deltacnn}
Parger, M., Tang, C., Twigg, C.~D., Keskin, C., Wang, R., and Steinberger, M.
\newblock Deltacnn: End-to-end cnn inference of sparse frame differences in
  videos.
\newblock \emph{CVPR 2022}, June 2022.

\bibitem[Radford et~al.(2019)Radford, Wu, Child, Luan, Amodei, and
  Sutskever]{GPT2}
Radford, A., Wu, J., Child, R., Luan, D., Amodei, D., and Sutskever, I.
\newblock Language models are unsupervised multitask learners.
\newblock 2019.

\bibitem[Ren et~al.(2018)Ren, Pokrovsky, Yang, and Urtasun]{ren2018sbnet}
Ren, M., Pokrovsky, A., Yang, B., and Urtasun, R.
\newblock Sbnet: Sparse blocks network for fast inference.
\newblock In \emph{Proceedings of the IEEE Conference on Computer Vision and
  Pattern Recognition}, pp.\  8711--8720, 2018.

\bibitem[Sanh et~al.(2020)Sanh, Debut, Chaumond, and Wolf]{sanh2020distilbert}
Sanh, V., Debut, L., Chaumond, J., and Wolf, T.
\newblock Distilbert, a distilled version of bert: smaller, faster, cheaper and
  lighter, 2020.

\bibitem[Sharir \& Shashua(2018)Sharir and Shashua]{sharir2018expressive}
Sharir, O. and Shashua, A.
\newblock On the expressive power of overlapping architectures of deep
  learning.
\newblock In \emph{6th International Conference on Learning Representations
  (ICLR)}, 2018.

\bibitem[Shi et~al.(2022)Shi, Zhao, Tang, Wang, Li, Bi, Jiang, Huang, Cui,
  Huang, Zhou, Dai, and Ma]{shi2022effidit}
Shi, S., Zhao, E., Tang, D., Wang, Y., Li, P., Bi, W., Jiang, H., Huang, G.,
  Cui, L., Huang, X., Zhou, C., Dai, Y., and Ma, D.
\newblock Effidit: Your ai writing assistant, 2022.

\bibitem[van~den Oord et~al.(2017)van~den Oord, Vinyals, and
  kavukcuoglu]{VQ2017_7a98af17}
van~den Oord, A., Vinyals, O., and kavukcuoglu, k.
\newblock Neural discrete representation learning.
\newblock In Guyon, I., Luxburg, U.~V., Bengio, S., Wallach, H., Fergus, R.,
  Vishwanathan, S., and Garnett, R. (eds.), \emph{Advances in Neural
  Information Processing Systems}, volume~30. Curran Associates, Inc., 2017.
\newblock URL
  \url{https://proceedings.neurips.cc/paper_files/paper/2017/file/7a98af17e63a0ac09ce2e96d03992fbc-Paper.pdf}.

\bibitem[Vaswani et~al.(2017)Vaswani, Shazeer, Parmar, Uszkoreit, Jones, Gomez,
  Kaiser, and Polosukhin]{transformers}
Vaswani, A., Shazeer, N., Parmar, N., Uszkoreit, J., Jones, L., Gomez, A.~N.,
  Kaiser, L.~u., and Polosukhin, I.
\newblock Attention is all you need.
\newblock In Guyon, I., Luxburg, U.~V., Bengio, S., Wallach, H., Fergus, R.,
  Vishwanathan, S., and Garnett, R. (eds.), \emph{Advances in Neural
  Information Processing Systems}, volume~30. Curran Associates, Inc., 2017.
\newblock URL
  \url{https://proceedings.neurips.cc/paper_files/paper/2017/file/3f5ee243547dee91fbd053c1c4a845aa-Paper.pdf}.

\bibitem[Xiao et~al.(2022)Xiao, Lin, Seznec, Wu, Demouth, and
  Han]{xiao2022smoothquant}
Xiao, G., Lin, J., Seznec, M., Wu, H., Demouth, J., and Han, S.
\newblock Smoothquant: Accurate and efficient post-training quantization for
  large language models.
\newblock \emph{arXiv}, 2022.

\bibitem[Yu et~al.(2022)Yu, Li, Koh, Zhang, Pang, Qin, Ku, Xu, Baldridge, and
  Wu]{yu2022vectorquantized}
Yu, J., Li, X., Koh, J.~Y., Zhang, H., Pang, R., Qin, J., Ku, A., Xu, Y.,
  Baldridge, J., and Wu, Y.
\newblock Vector-quantized image modeling with improved {VQGAN}.
\newblock In \emph{International Conference on Learning Representations}, 2022.
\newblock URL \url{https://openreview.net/forum?id=pfNyExj7z2}.

\bibitem[Yu et~al.(2017)Yu, Liu, Wang, and Tao]{Yu_2017_CVPR}
Yu, X., Liu, T., Wang, X., and Tao, D.
\newblock On compressing deep models by low rank and sparse decomposition.
\newblock In \emph{Proceedings of the IEEE Conference on Computer Vision and
  Pattern Recognition (CVPR)}, July 2017.

\bibitem[Zeghidour et~al.(2021)Zeghidour, Luebs, Omran, Skoglund, and
  Tagliasacchi]{zeghidour2021soundstream}
Zeghidour, N., Luebs, A., Omran, A., Skoglund, J., and Tagliasacchi, M.
\newblock Soundstream: An end-to-end neural audio codec, 2021.

\bibitem[Zhang et~al.(2022)Zhang, Roller, Goyal, Artetxe, Chen, Chen, Dewan,
  Diab, Li, Lin, Mihaylov, Ott, Shleifer, Shuster, Simig, Koura, Sridhar, Wang,
  and Zettlemoyer]{zhang2022opt}
Zhang, S., Roller, S., Goyal, N., Artetxe, M., Chen, M., Chen, S., Dewan, C.,
  Diab, M., Li, X., Lin, X.~V., Mihaylov, T., Ott, M., Shleifer, S., Shuster,
  K., Simig, D., Koura, P.~S., Sridhar, A., Wang, T., and Zettlemoyer, L.
\newblock Opt: Open pre-trained transformer language models, 2022.

\end{thebibliography}
\bibliographystyle{icml2023}

\newpage
\appendix
\onecolumn

\section{Incremental Computation of Other Operations}\label{app:other-ops}

\subsection{Efficient Self-Attention}
After per-location operations, the next most intensive operation is the self-attention mechanism at the core of the transformers network.
Recall that self-attention takes a feature-map tensor $X$, and first applies on it per-location linear projections to form query, key, and value tensors, denoted by $Q , K \in \R^{B \times N \times d_{QK}}$ and $V \in \R^{B \times N \times d_{V}}$, respectively.
These tensors are combined to form the output, denoted by $O \in \R^{B \times N \times d_V}$, as follows:
\begin{align}\label{eq:attention}
O_{bnk} &= \sum_{i=1}^N \overbrace{\sigma\left(\sum_{j=1}^{d_{QK}} Q_{bnj} K_{bij}	\right)}^{\textrm{``Attention matrix''}} V_{bik},
\end{align}
where the motivation behind it is that the \emph{attention matrix} term (matrix per batch position) represents the relevance of the key at sequence position $i$ to the query at position $n$, according to which each of the value vectors is aggregated per batch and sequence position.
In practice, multiple self-attention ``heads'' are performed in parallel and their output is concatenated along the vector (3rd) dimension, followed by another linear projection.
Additionally, for autoregressive networks, the attention matrix is also ``masked'' (multiplied with zeros) where $n \geq j$ to ensure the network's outputs form valid conditional distributions.

While many of the internal operations in the self-attention module falls under the efficient per-location operations we already covered, the core part described by eq.~\ref{eq:attention} does not and so requires a different approach.
The general approach here would be to first compute the full attention matrix and outputs with respect to the base set of indices.
Now, for any batch position, we have the associated sequence locations that are different from the base indices, as well as their respective values.
Observe the effect these conflicting keys and queries have on their attention matrix relative to the base's attention matrix~--~each conflicting location contributes to a single altered row and altered column associated with a modified query and key, respectively.
Based on our assumption that each batch position has a constant number of conflicting locations, it means we only have to recompute $O(N)$ entries in the $N$-by-$N$ attention matrix for each batch position, while the rest can be assumed unchanged from the base attention matrix.

Next, to compute the final outputs, we do not need to compute the entire outer sum in eq.~\ref{eq:attention} for every batch and sequence position.
Every coordinate $(b,n)$ that is different from the base indices, corresponds to an altered query location and so we have the entire respective row in its attention matrix and the corresponding values from $V$.
For the other coordinates, we only need to correct the output values precomputed for the base for the specific columns in the attention matrix that were affected by the conflicting keys and values.
We can do so by subtracting at the corresponding locations the  multiplication of the base attention matrix and values and adding the altered attention matrix entries and altered values.
In total, this amounts to $O(BN(d_{QK} + d_V))$ operations, as there are $O(B)$ batch-sequence locations that require $O(Nd_{QK} + Nd_V)$ operations (compute row in attention matrix and multiplication per value), and the other locations, bounded by $O(BN)$ that requires just $O(d_{QK} + d_V)$ (compute a few altered keys and values).

\subsection{Efficient Vector Quantization}

Notice that the output of self-attention is dense and can no longer be represented in the compressed form.
The role of the VQ operation following self-attention is to ensure the required properties of our compressed format hold throughout the neural network.
However, VQ itself has a cost of comparing each vector in batch-sequence position to its codebook, equivalent to $O(Qd_V^2)$ per location~--~the cost of computing the Euclidean distance for every codebook vector.
Notice that it is a per-location operation, and while the input is not compressed, we can leverage another trick.
The euclidean distance can be rewritten as $\norm{\x - \q}^2 = \norm{\x}^2 - 2\x^T\q + \norm{\q}^2$, and so the minimizer of $\min_i \norm{\x - \q^i}$ is equivalent to the maximizer of $\min_i \x^T\q^i + b_i$ where $b_i = - \nicefrac{\norm{\q^i}^2}{2}$ can be computed once for all input vectors.
Next, observe that the self-attention operation is a \emph{linear} operation with respect to the values tensor.
This means that the inner product between every output vector of the attention matrix and every codebook vector is equivalent to computing the inner products on the \emph{compressed} values tensor and then applying the attention matrix to the result.
Given that it is a per-location operation, it is efficient and so we can effectively ``hide'' the cost of VQ itself.

\subsection{Efficient Binary Element-wise Operations}

The final kind of operation we must realize is any binary element-wise operation between two compressed feature maps, i.e., $F(X, Y)_{ijk} = f(X_{ijk}, Y_{ijk})$ where $f$ is addition, multiplication, etc.
If the two compressed tensors share the same indices matrix, then the operation again reduces to operating on their respective codebooks, similarly to the per-location operations.
Otherwise, we can leverage the fact each tensor is composed of few repeating vectors, and so perform the underlying binary operation $f$ on just the unique combinations of vectors, which results in a both a new codebook and a new indices matrices.
We can perform the above in two steps: (i) Find the unique pairs of elements in the two indices matrices that correspond to the unique pairs of vectors, which due to its sparse representation results in a $O(B \log_2 B)$ complexity; and (ii) apply the underlying operation $f$ on the pairs of vectors from each respective codebook corresponding to the unique pairs from the first step.
Naturally, in the worst case of two completely incompatible compressed feature maps, such an operation could result in doubling the size of the codebook.
However, since we assume both compressed feature maps are linked to the same input with its inherent self-similarity, then the two feature maps will agree on most indices and so we can expect an additive, rather than multiplicative, increase in the codebook size.
Thus, the cost of such operations will be on the order of $O(B \log_2 B + Qd)$.
For $B = O(N)$ and $Q = O(N)$, this amounts to $O(N(d + \log_2 N))$. For any reasonable choice of $d$ (i.e., $d \geq 20$), then for all practical purposes, we can assume $d \gg \log_2 N$, and so we can simply write $O(Nd)$, and again observe a theoretical speedup of $O(1/N)$.

\section{Training with Large Sampled Positional Embeddings}\label{app:coupon}

We can use such a large number because all positions will be encountered very frequently during training.
Per the well-known coupons collector problem, if we have $k$ positional embeddings and we sample each independently, it will take on average $k \log k$ steps to sample all positions\footnote{This is an overestimate, as in practice we sample in groups without replacement.}.
Assuming the maximum training length is 2048, $k = 100 \cdot 2048$, and a modest batch size of 8, then every 220 iterations all positions will be seen, making training such large positional embeddings practical.

\end{document}